\documentclass{article} 
\usepackage[preprint]{colm2026_conference}

\usepackage{microtype}
\usepackage{hyperref}
\usepackage{url}
\usepackage{booktabs}
\usepackage{amsmath}
\usepackage{amssymb}
\usepackage{graphicx}
\usepackage{multirow}
\usepackage{float}
\usepackage{caption}
\usepackage{enumitem}

\usepackage{lineno}

\newcommand{\SL}{\textsc{sl}}
\newcommand{\PS}{\textsc{ps}}
\newcommand{\CL}{\textsc{cl}}
\newcommand{\SSI}{\ensuremath{\text{SSI}}}

\definecolor{darkblue}{rgb}{0, 0, 0.5}
\hypersetup{colorlinks=true, citecolor=darkblue, linkcolor=darkblue, urlcolor=darkblue}

\title{Polysemanticity or Polysemy? Lexical Identity Confounds Superposition Metrics}

\author{Iyad Ait Hou \\
Department of Computer Science\\
George Washington University\\
Washington, D.C, USA \\
\texttt{{iyad.aithou@gwu.edu}} \\
\And
Rebecca Hwa \\
Department of Computer Science \\
George Washington University \\
Washington, D.C, USA \\
\texttt{rebecca.hwa@gwu.edu} \\
\AND
}

\begin{document}

\ifcolmsubmission
\linenumbers
\fi

\maketitle

\begin{abstract}
If the same neuron activates for both ``lender'' and ``riverside,''
standard metrics attribute the overlap to superposition---the neuron
must be compressing two unrelated concepts. This work explores how much of the overlap is
due to a lexical confound: neurons fire for a shared \emph{word form} (such as ``bank'') rather than 
for two compressed concepts.
A $2 \times 2$ factorial decomposition reveals that the lexical-only
condition (same word, different meaning) consistently exceeds the
semantic-only condition (different word, same meaning) across models
spanning 110M--70B parameters.
The confound carries into sparse autoencoders (18--36\% of features
blend senses), sits in $\le$\,1\% of activation dimensions,
and hurts downstream tasks: filtering it out improves word sense
disambiguation and makes knowledge edits more selective ($p = 0.002$).
\end{abstract}
\section{Introduction}

Polysemanticity~\citep{elhage2022superposition, olah2020zoom} describes the phenomenon in which a neuron in a large language model is activated for multiple, seemingly unrelated concepts. An explanation for this is superposition: when a model lacks the dimensions to represent all its learned concepts separately, it packs more variables into its activation space than it has neurons. By storing these variables as nearly-orthogonal directions, the model can function despite a single neuron participating in several distinct concepts~\citep{elhage2022superposition}. In this light, polysemanticity looks like a compression artifact, and sparse autoencoders (SAEs) are trained to undo it~\citep{bricken2023monosemanticity, templeton2024scaling}.

However, some cases of polysemanticity may be a lexical confound rather than genuine compression. A neuron activating for both ``financial bank'' and ``river bank'' is labeled polysemantic~\citep{olah2020zoom, bricken2023monosemanticity}, but the overlap may also reflect the model processing a shared word form before disambiguating its meaning~\citep{ethayarajh2019contextual}. Standard metrics cannot tell these apart, and so they risk overestimating superposition-driven variable sharing.

We therefore ask: \emph{to what extent are current measurements of polysemanticity actually capturing lexical identity rather than superposition?} We hypothesize that much of the measured overlap is driven by shared word forms rather than the need to compress unrelated concepts, and that these lexical signals play a primary, causal role in the model's processing. To test this, we isolate the drivers of activation overlap using a $2 \times 2$ factorial decomposition (Figure~\ref{tab:intro_design}) that pairs sentences along two axes---same/different word and same/different meaning. The lexical-only condition (same word, different meaning) can then be compared directly against the semantic-only condition (different word, same meaning): whichever produces more overlap reveals the main driver.

\begin{figure}[t]
\centering
\includegraphics[width=\columnwidth]{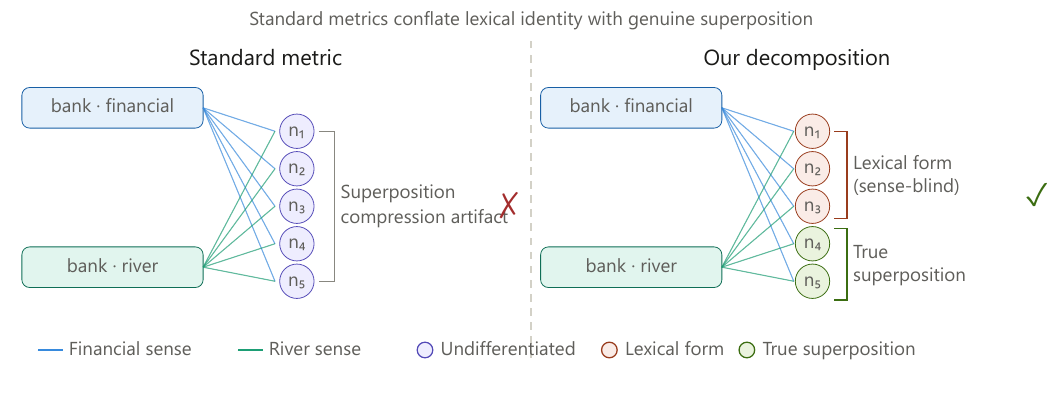}
\caption{\textbf{Overview.} Standard metrics (top) see that neurons $n_1$--$n_5$ fire for both senses of \emph{bank} and call all of it superposition. Our decomposition (bottom) shows that most shared neurons are \emph{sense-blind}---they encode the word form, not compressed concepts. Only a small remainder is genuine superposition.}
\label{fig:overview}
\end{figure}

The results show a consistent pattern: word form wins. Across nine transformers spanning 110M to 70B parameters, four architecture families, and 407 polysemous words, the lexical-only condition consistently exceeds the semantic-only condition (Figure~\ref{fig:layer_trends}). The confound carries into SAEs, where 18--36\% of learned features blend both senses of a word. It sits in a compact 20-dimensional subspace ($\le$\,1\% of activation dimensions) that can be cleanly removed. And it matters in practice: sense-blind neurons hurt word sense disambiguation (6~pp below sense-selective neurons) and make ROME knowledge edits less selective ($p = 0.002$). Figure~\ref{fig:overview} shows the core idea.

\section{Related work}
\label{sec:related}

\paragraph{Polysemanticity and superposition.}
The superposition hypothesis~\citep{elhage2022superposition} frames
polysemanticity as a compression artifact, and SAEs are designed to
recover monosemantic features from it~\citep{bricken2023monosemanticity,
templeton2024scaling, sharkey2022taking}.
Recent work has shown that polysemanticity can arise even without capacity
pressure~\citep{lecomte2025incidental} and has tracked its
emergence during training~\citep{wu2025tokens}.
None of these works examine whether the \emph{inputs} triggering
multi-concept activation share a lexical form---the confound we identify.

\paragraph{Word sense representations.}
\citet{ethayarajh2019contextual} and \citet{wiedemann2019does} established
that contextual embeddings cluster by word identity before sense.
This is a geometric observation about full hidden states; it was not
connected to polysemanticity metrics or to individual neurons.
Our work bridges this gap: we operate at the level of \emph{individual
MLP neurons} (the unit at which polysemanticity is defined), provide a
\emph{quantitative metric} ($R_\text{lex}$), and trace the confound
through \emph{causal and downstream consequences}.
\citet{lyu2025homonymy} and \citet{minegishi2025rethinking} study polysemous
words in neural representations but do not measure how the confound
affects interpretability tools.

\paragraph{Causal methods and model editing.}
Probing classifiers~\citep{conneau2018probing} test what information a
representation encodes; activation patching~\citep{vig2020investigating,
meng2022locating} and mean-ablation~\citep{chan2022causal} test whether
that information plays a causal role.
We use both to validate the sense-selective / sense-blind distinction.
Model editing methods such as ROME~\citep{meng2022locating} and
embedding-level interventions~\citep{aithou2026parameter} modify
specific representations to change model behavior; none account for
lexical identity as a source of collateral damage across word senses.

\section{Methodology}
\label{sec:method}

Our approach has three parts: (1)~a factorial decomposition that separates lexical from semantic contributions to neuron overlap, (2)~a neuron classification that identifies which neurons encode word form vs.\ word meaning, and (3)~causal and downstream experiments that test whether the distinction matters. All analyses target MLP intermediate activations, the level at which polysemanticity is typically analyzed~\citep{olah2020zoom, elhage2022superposition, bricken2023monosemanticity}.

\subsection{Decomposition framework}
\label{sec:design}

\begin{figure}[t]
\centering
\includegraphics[width=0.75\columnwidth]{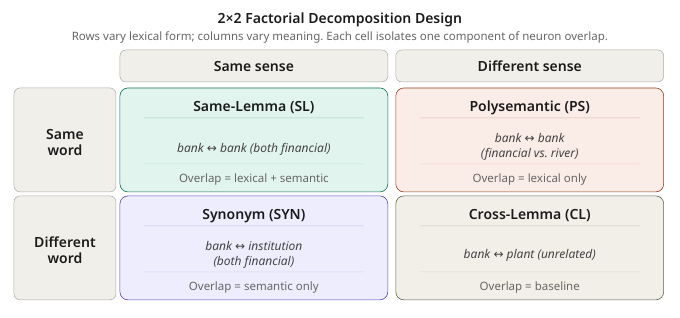}
\caption{$2 \times 2$ factorial decomposition design.}
\label{tab:intro_design}
\end{figure}

We construct sentence pairs crossing two binary factors---same/different word and same/different sense---yielding four conditions (Figure~\ref{tab:intro_design}):
\textbf{SL}~(same word, same sense) produces the most neuron overlap because both word form and meaning are shared;
\textbf{CL}~(different word, different sense) produces the least because neither is shared.
The two off-diagonal conditions isolate the drivers:
\textbf{PS}~(same word, different sense) measures overlap from shared word form alone, and
\textbf{SYN}~(different word, same sense) measures overlap from shared meaning alone.

For each sentence, we extract MLP intermediate activations at the last subword of the target word~\citep{ethayarajh2019contextual} and compute pairwise overlap via three metrics: cosine similarity, Jaccard overlap of active-neuron sets, and magnitude divergence, aggregated per word with bootstrap 95\% CIs.

\paragraph{Lexical contribution ratio.}
If lexical and semantic contributions do not interact (validated in Section~\ref{sec:results-controls}), their relative sizes can be read off directly. Normalizing by the total range gives the \emph{lexical contribution ratio}:
\begin{equation}
  R_\text{lex} = \frac{\bar{M}_\text{Polysemy} - \bar{M}_\text{Synonym}}
                      {\bar{M}_\text{Same-Lemma} - \bar{M}_\text{Cross-Lemma}}
  \label{eq:rlex}
\end{equation}
The numerator is the difference between same-word overlap and same-meaning overlap---the excess due to word form. The denominator is the full range from maximum to minimum overlap. $R_\text{lex} > 0$ means word form contributes more overlap than meaning; $R_\text{lex} = 1$ means word form accounts for all of it.
As a control, we also compute $R_\text{lex}$ from raw token embeddings (before any transformer computation); MLP-level $R_\text{lex}$ exceeding this value indicates the confound is not simply inherited from the input encoding.

\subsection{Neuron classification}
\label{sec:neuron-class}

The decomposition above measures overlap at the population level.
To test whether individual neurons behave differently depending on
whether they encode word form or word meaning, we classify them
into two groups.

\paragraph{Sense Selectivity Index (SSI).}
For each polysemous word, we have sentences for two distinct senses (e.g., ``bank'' as financial institution = sense~A, ``bank'' as riverbank = sense~B). For each neuron $j$, we compute Cohen's $d$ between its activations on sense-A vs.\ sense-B sentences:
\begin{equation}
  \SSI_j = \frac{|\mu_A^j - \mu_B^j|}{\sigma_\text{pooled}^j}
  \label{eq:ssi}
\end{equation}
Neurons with $\SSI > 2$ are \emph{sense-selective}; neurons with
$\SSI < 0.5$ (above median activation) are \emph{sense-blind}.
These thresholds follow standard conventions for Cohen's $d$ (large
vs.\ small effect). The condition ordering (PS~$>$~SYN) and the
sense-blind/selective dissociation do not depend on the exact threshold
choice; SSI distributions are reported in Appendix~\ref{app:ssi}.

\paragraph{Form detectors (sense-label-free).}
To avoid circularity, we independently identify form detectors using only
word identity: for each word $w$ and neuron $j$, we compute
\emph{consistency} ($1 - \text{CV}$ of activations across all $w$-sentences)
and \emph{specificity} (Cohen's $d$ vs.\ other words).
Neurons ranking in the top $K$ by consistency $\times$ specificity
are form detectors for $w$---they fire reliably for $w$ regardless of sense.

\subsection{Causal intervention protocol}
\label{sec:intervention}

If sense-selective and sense-blind neurons truly encode different information,
removing one group should affect model behavior differently from removing the
other. We test this via mean-ablation~\citep{chan2022causal}: replacing selected
neurons' activations with their dataset mean, simultaneously across all layers
at the target-word position.
Neuron counts are matched across groups (sense-A-selective, sense-B-selective,
sense-blind, random) at each layer.
We evaluate via:
\textbf{(1)} KL divergence of the full output distribution,
\textbf{(2)} sense accuracy (diagnostic token probabilities), and
\textbf{(3)} sense-specific perplexity change, with
specificity $= |\Delta\text{ppl}_\text{target} -
\Delta\text{ppl}_\text{other}|$.

\section{Experimental setup}
\label{sec:setup}

\paragraph{Dataset.}
We use SemCor~\citep{miller1993semantic}, the only large-scale corpus with
human-annotated WordNet sense tags (37{,}176 sentences).
From it we select 407 content words (nouns/verbs, $\geq 2$ senses,
$\geq 5$ sentences per sense, Wu-Palmer similarity $< 0.50$ between senses; varying this threshold from 0.35 to 0.65 does not change the condition ordering).
SYN pairs use WordNet synonyms attested in SemCor; CL pairs sample unrelated
words. Results replicate across nine models trained on four corpora and
validate on modern Wikipedia text (Appendix~\ref{app:modern},
\ref{app:syn_validation}, \ref{app:dataset}).

\paragraph{Models.}
We evaluate nine transformers spanning four architecture families and 110M--70B parameters (Table~\ref{tab:models}). Models up to 13B are loaded in \texttt{float32}; LLaMA-2-70B uses NF4 quantization.

\begin{table}[t]
\begin{center}
\small
\setlength{\tabcolsep}{3pt}
\begin{tabular}{lrrrll}
\toprule
\textbf{Model} & \textbf{Params} & \textbf{Layers} & $d_\text{mlp}$ & \textbf{Type} & \textbf{Ref.} \\
\midrule
GPT-2         & 117M & 12 & 3072  & Auto & \citealt{radford2019language} \\
GPT-2-Med     & 345M & 24 & 4096  & Auto & \citealt{radford2019language} \\
BERT-base     & 110M & 12 & 3072  & Bidir & \citealt{devlin2019bert} \\
ELECTRA-base  & 110M & 12 & 3072  & Bidir & \citealt{clark2020electra} \\
Pythia-1B     & 1.0B & 16 & 8192  & Auto & \citealt{biderman2023pythia} \\
Pythia-6.9B   & 6.9B & 32 & 16384 & Auto & \citealt{biderman2023pythia} \\
Pythia-12B    & 12B  & 36 & 20480 & Auto & \citealt{biderman2023pythia} \\
LLaMA-2-13B  & 13B  & 40 & 13824 & Auto & \citealt{touvron2023llama} \\
LLaMA-2-70B  & 70B  & 80 & 28672 & Auto & \citealt{touvron2023llama} \\
\bottomrule
\end{tabular}
\end{center}
\caption{Models evaluated.  Auto = autoregressive LM; Bidir = bidirectional (masked LM / replaced-token detection).  LLaMA-2-70B uses NF4 quantization.}
\label{tab:models}
\end{table}

\section{Results}
\label{sec:results}

We present the evidence in stages: the confound appears consistently across layers (Section~\ref{sec:results-layers}), replicates across model scale (Section~\ref{sec:results-cross}) and into SAEs (Section~\ref{sec:results-sae}), has identifiable mechanistic structure (Sections~\ref{sec:results-form}--\ref{sec:results-lis}), and affects downstream tasks (Section~\ref{sec:results-rome}). Primary results use GPT-2; replication across models follows in Section~\ref{sec:results-cross}.

\subsection{The lexical confound across layers}
\label{sec:results-layers}

\begin{figure}[t]
\centering
\includegraphics[width=\columnwidth]{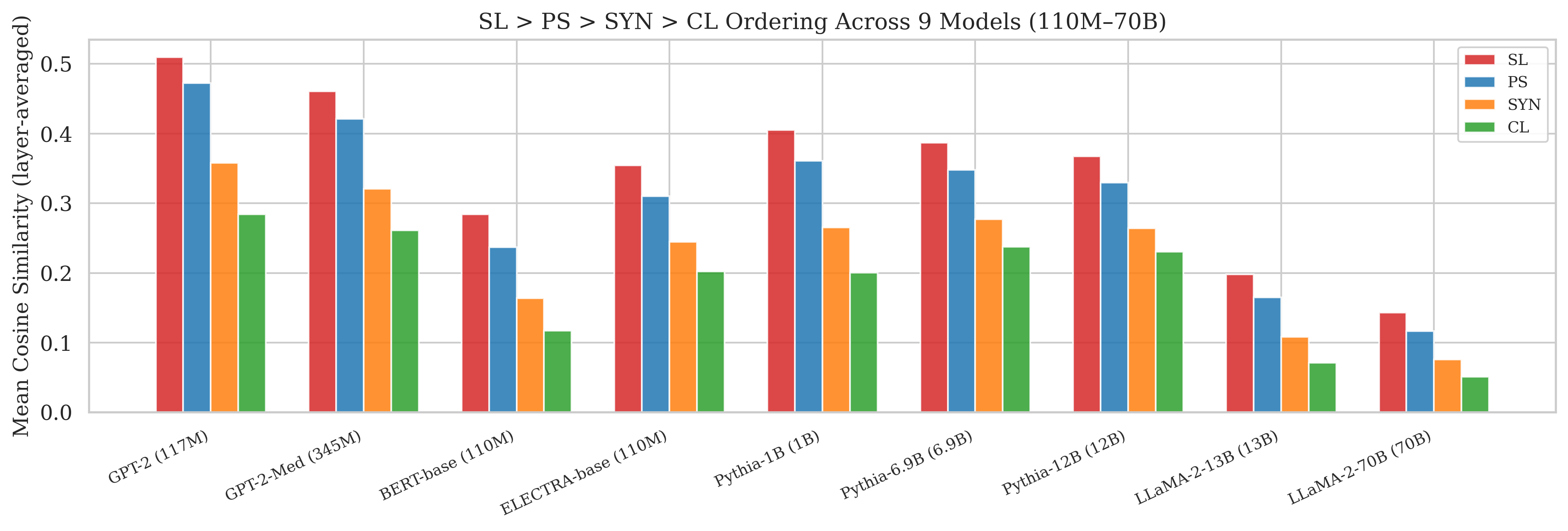}
\vspace{-1mm}
\includegraphics[width=\columnwidth]{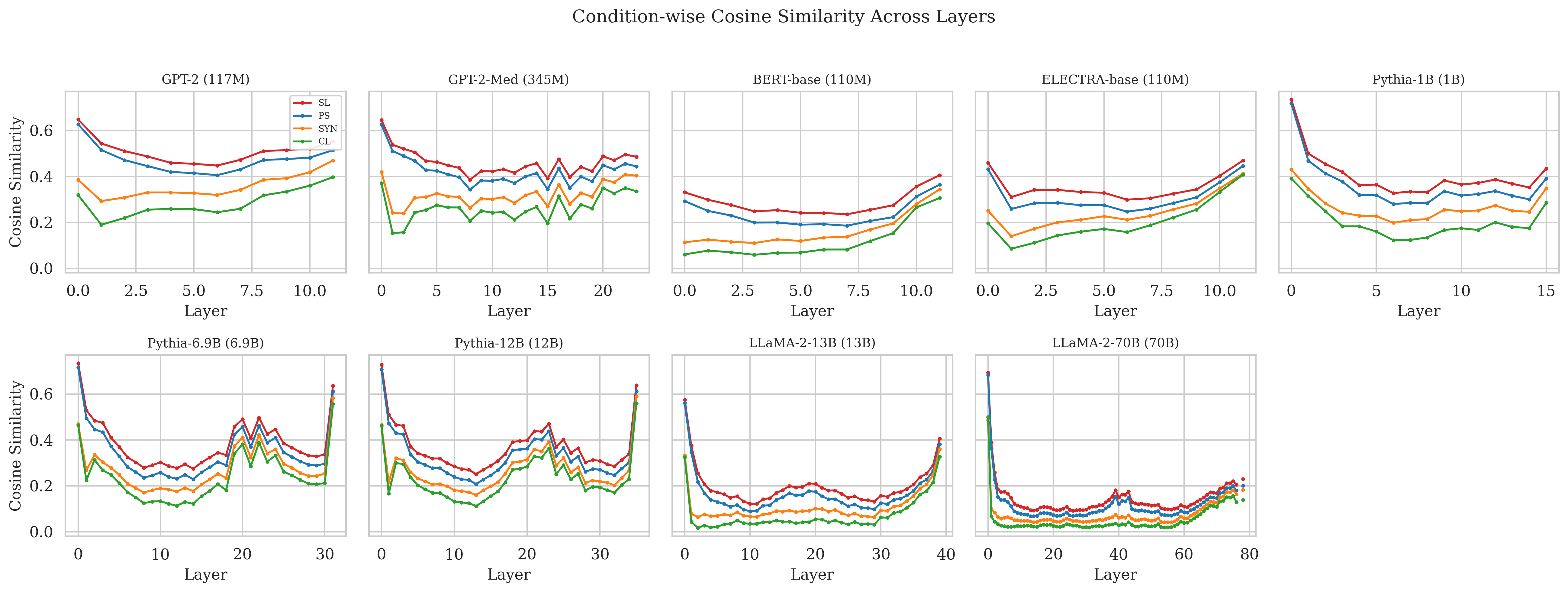}
\vspace{-3mm}
\caption{\textbf{Top}: Layer-averaged cosine similarity by condition across nine models (110M--70B). PS consistently exceeds SYN: word form drives more overlap than shared meaning.
\textbf{Bottom}: Per-layer breakdown showing the pattern holds across layers. $R_\text{lex}$ trends in Appendix~\ref{app:results}.}
\label{fig:layer_trends}
\end{figure}

Cosine similarity follows $\SL > \PS > \text{SYN} > \CL$ across layers and models (Figure~\ref{fig:layer_trends}); the same ordering holds for Jaccard overlap and magnitude divergence (Appendix~\ref{app:figures}). Crucially, PS consistently exceeds SYN (Wilcoxon $p < 0.001$, Holm-Bonferroni corrected): sharing a word form produces more overlap than sharing a meaning. A token-embedding baseline ($R_\text{lex}^\text{emb} = 0.71$--$0.88$) confirms the confound is not merely inherited from token embeddings (Table~\ref{tab:emb_baseline}).

$R_\text{lex}$ decreases with depth (0.74 to 0.29 across GPT-2 layers; Figure~\ref{fig:rlex_app}). In all models tested, the bootstrap CI remains above zero at all layers---though we note this is an empirical observation, not a guaranteed property; a model with stronger disambiguation could in principle drive $R_\text{lex}$ to zero or below. Sense-selective neurons ($\SSI > 2$) stay below $1\%$ of the population, confirming sense information is encoded by a very sparse subpopulation.

\subsection{Cross-model consistency}
\label{sec:results-cross}

The PS~$>$~SYN gap replicates across all nine models tested, from
110M to 70B parameters (Figure~\ref{fig:layer_trends};
Appendix~\ref{app:results}). Bidirectional models (BERT, ELECTRA)
show steeper $R_\text{lex}$ declines with depth; autoregressive models
maintain the pattern through 12--13B. LLaMA-2-70B (NF4 quantization,
80 layers) shows the same ordering (layer-averaged: $\PS = .117$
vs.\ $\text{SYN} = .076$; $R_\text{lex} = 0.45$). The neuron
classification and causal experiments use models up to 13B in full
precision; the 70B result confirms the condition ordering extends
to larger scale.

\subsection{Impact on SAEs and standard metrics}
\label{sec:results-sae}

\begin{figure}[t]
\centering
\includegraphics[width=\columnwidth]{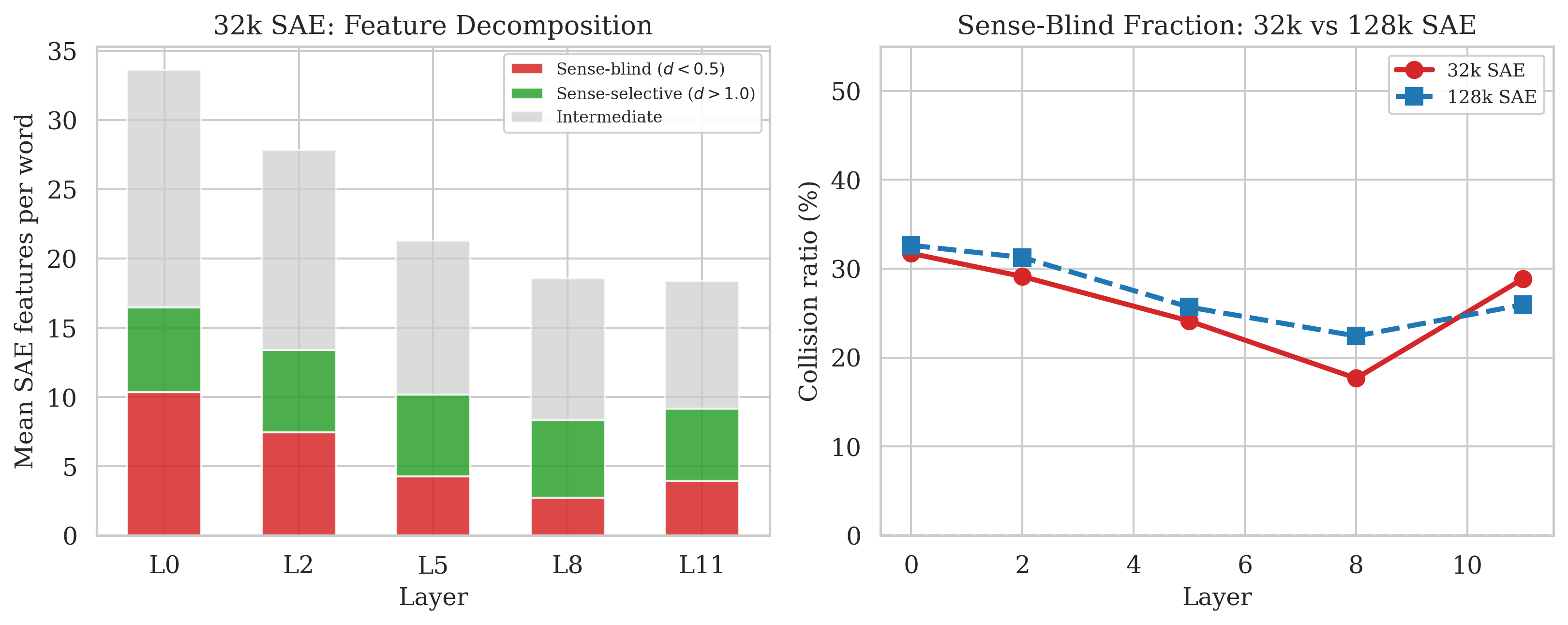}
\caption{SAE collision analysis (GPT-2). \textbf{Left}: mean features per word by
sense discriminability.  \textbf{Right}: collision ratio across layers.
18--32\% of features conflate senses of the same word (18--36\% including Pythia-410M; Section~\ref{sec:results-sae}).}
\label{fig:sae_collision}
\end{figure}

Pre-trained GPT-2 SAEs~\citep{bloom2024saetraining} (32k and 128k features)
inherit the confound: 18--32\% of active features per layer are sense-blind,
and the 128k SAE ($4\times$ capacity) produces nearly identical ratios
(Figure~\ref{fig:sae_collision}). On Pythia-410M (65k SAEs), ratios are
19--36\%. SAEs do \emph{not} resolve the lexical confound.

\paragraph{Individual feature analysis.}
Inspecting individual sense-blind features reveals the scope of the problem.
Feature \#5132 in the 32k GPT-2 SAE at layer~8 (the OAI v5 release;
publicly inspectable on Neuronpedia\footnote{\url{https://www.neuronpedia.org/gpt2-small/8-res_post_32k-oai/5132}})
fires with similar activation (Cohen's $d < 0.5$) for \textbf{232 different
polysemous words}, spanning unrelated domains (\emph{foot}, \emph{church},
\emph{kick}, \emph{dance}, \emph{win}, \emph{cell}).
Its auto-generated label on Neuronpedia is ``intensifying adjectives and
adverbs describing sound or scale,'' yet it fires on 47.5\% of all inputs---a
density incompatible with a narrow semantic category.
This feature encodes lexical form, not a coherent concept; its auto-interp label reflects surface patterns rather than its actual function.

Applying standard polysemanticity scores to our dataset, $\approx\!57\%$ of activating input pairs for flagged neurons (conditional on polysemous inputs) share a word form rather than a genuinely different concept. This is not a global inflation rate---but polysemous words are common ($>$80\% of frequent English vocabulary; \citealt{rodd2002making}).

\subsection{Neuron-level validation}
\label{sec:results-form}

The preceding sections establish the confound at the population level. We now ask whether it holds at the level of individual neurons.

\paragraph{Form detectors are sense-blind.}
To validate the confound non-circularly, we identify form detectors using only
word identity (Section~\ref{sec:neuron-class}) and separately classify them
by SSI.
On GPT-2, $67.3\%$ of form detectors are sense-blind ($\SSI < 0.5$), while
only $0.1\%$ are sense-selective---establishing that the neurons
polysemanticity metrics flag are predominantly lexical identity detectors.

\paragraph{Probing confirms the functional split.}
\label{sec:results-probe}
Logistic regression probes (LOO CV, $C = 0.01$) on neuron-group activations
reveal a clear dissociation: selective neurons achieve $91.1\%$ sense
accuracy; blind neurons $41.9\%$ (near chance); random $68.9\%$.
For word-form detection, all groups perform comparably ($\sim$85\%),
confirming word identity is distributed.
Blind neurons encode form (85\%) but not sense (42\%); selective neurons
encode both.

\subsection{Causal intervention: sense-specific editing}
\label{sec:results-causal}

We test whether the distinction has \emph{causal} consequences via mean-ablation. Ablating sense-selective neurons produces $3$--$5\times$ larger KL divergence than matched sense-blind or random groups (up to 345M; $1.2$--$2.1\times$ at 12--13B), suggesting sense-selective neurons carry disproportionate causal weight despite small absolute effects ($10^{-3}$--$10^{-4}$ KL, as expected when ablating $\sim$50 of 3k--20k neurons).

\paragraph{Sense-specific editing.}
Ablating across all layers with matched neuron counts
(Table~\ref{tab:editing}), sense-selective ablation is $6.6\times$ more
specific than blind on GPT-2 (53 words).
Sense-A ablation raises perplexity on sense-A sentences ($+0.55$) while
leaving sense-B largely unchanged ($-0.10$); blind ablation has minimal effect
on either sense ($-0.07$, $+0.17$).

\begin{table}[t]
\begin{center}
\small
\begin{tabular}{lccl}
\toprule
\textbf{Ablation} & \textbf{$\Delta$ppl$_A$} & \textbf{$\Delta$ppl$_B$} & \textbf{Specificity} \\
\midrule
Sense-A neurons & $+0.55$ & $-0.10$ & $0.66$ \\
Sense-B neurons & $+0.10$ & $+2.62$ & $2.52$ \\
Sense-blind     & $-0.07$ & $+0.17$ & $0.24$ \\
Random          & $-0.64$ & $-0.32$ & $0.32$ \\
\bottomrule
\end{tabular}
\end{center}
\caption{Sense-specific editing (GPT-2, 53 words, 142 neurons per group).
Sense-selective ablation targets one sense ($6.6\times$ specificity gap over blind).
Specificity $= |\Delta\text{ppl}_\text{target} - \Delta\text{ppl}_\text{other}|$.}
\label{tab:editing}
\end{table}

\paragraph{Collateral damage on individual words.}
Per-word ablation at layer~6 confirms the pattern: for \emph{match}, sense-blind ablation damages both senses equally (specificity $= 0.010$), while sense-selective ablation is $5.5\times$ more targeted; for \emph{light}, $12\times$. Sense-blind neurons cannot be used to target one meaning without equally affecting the other.

\subsection{Controls}
\label{sec:results-controls}

The interaction term $I$ is small in all models ($|I| < 0.04$), supporting
additivity. $R_\text{lex}$ correlates weakly with frequency ($r = 0.15$)
and non-significantly with position. A synonym-free variant and subset
invariance tests confirm robustness (Appendix~\ref{app:results},
\ref{app:syn_validation}; embedding baseline and interaction plots in
Appendix~\ref{app:figures}).

\subsection{The confound is correctable}
\label{sec:results-corrective}

The confound is not merely detectable---it is correctable.
We define a \emph{lexically-adjusted polysemanticity score}:
\begin{equation}
  P^\text{adj}_j = P^\text{raw}_j - \hat{\lambda}_\ell \cdot F_j
  \label{eq:adjusted}
\end{equation}
where $P^\text{raw}_j$ is the standard score, $F_j$ indicates whether neuron
$j$ is a form detector, and
$\hat{\lambda}_\ell = R_\text{lex}(\ell) \cdot \bar{P}_\ell$ is the
layer-specific expected lexical inflation.
On GPT-2, this reclassifies 6.5\% (layer~6) to 10.4\% (layer-averaged)
of neurons flagged as polysemantic, with 93--94\% of reclassified neurons
confirmed as form detectors ($\SSI < 0.5$).
The correction targets the top-50 form detectors per word ($\approx$1.6\% of GPT-2's 3072 MLP neurons), chosen to balance coverage with precision; this is a conservative lower bound.
This demonstrates \emph{correctability}: the confound can be identified and
removed from existing metrics.
The current implementation requires sense labels, limiting it to controlled
evaluations; extending this to an unsupervised correction is an open problem
(Section~\ref{sec:discussion}).

\subsection{Mechanistic analysis: lexical identity subspace}
\label{sec:results-lis}

Does the confound have specific geometric structure, or is it diffuse?
For each word $w_i$ with synonym data ($n = 62$), we compute
$\mathbf{d}_i = \bar{\mathbf{a}}_{w_i} - \bar{\mathbf{a}}_{\text{syn}_i}$
(mean activation minus synonym mean), isolating word form from meaning.
PCA on these difference vectors yields a \textbf{lexical identity subspace}
(LIS)~\citep[cf.][]{arditi2024refusal}. Removing just 20 LIS dimensions
($\le$\,1\% of activation space) reduces the PS--SYN gap by 37--51\%;
at 50 dimensions the gap reverses entirely (full dose-response in
Appendix~\ref{app:supplementary}). The confound is compact, localizable,
and consistent across GPT-2 and Pythia-1B.

The corrective score (Section~\ref{sec:results-corrective}) and the LIS show that the confound is not just measurable but removable. We now test whether doing so helps in practice.

\subsection{Downstream applications}
\label{sec:results-wsd}\label{sec:results-rome}

\begin{figure}[t]
\centering
\includegraphics[width=\columnwidth]{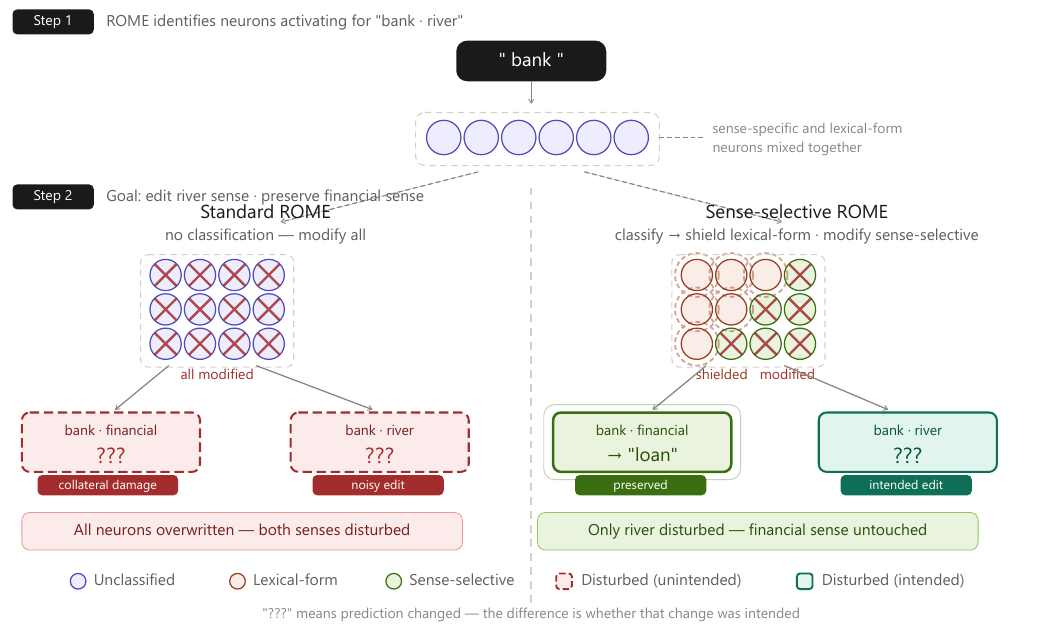}
\caption{Sense-selective ROME editing. Standard ROME modifies all activated neurons, disturbing both senses. Our approach classifies neurons first, shields sense-blind (lexical-form) neurons from modification, and edits only sense-selective ones---preserving the unedited sense.}
\label{fig:rome_panel}
\end{figure}

Finally, we demonstrate that the sense-selective / sense-blind distinction has practical value beyond diagnosis (Figure~\ref{fig:rome_panel}).

\paragraph{Word sense disambiguation.}
For each of 20 CoarseWSD-20 words, we train a logistic regression classifier
(5-fold CV) on MLP activations.
Sense-selective neurons (top 25\% by selectivity) achieve \textbf{89.8\%}
accuracy in GPT-2 (\textbf{89.3\%} in Pythia-1B), outperforming all neurons
(88.3\%) and sense-blind neurons (83.6\%/84.6\%).
The selective--blind gap (5--6~pp) peaks at early-to-mid layers.

\paragraph{ROME knowledge editing.}
We apply ROME~\citep{meng2022locating} to change one sense's prediction and
measure collateral damage on the other, comparing standard (all neurons),
filtered (sense-selective only), and blind-only edits.
Filtered edits are significantly more selective ($0.540$ vs.\ $0.519$,
$p = 0.002$, Wilcoxon; largest gain at layer~7: $+0.046$).
Blind-only edits are worst ($0.512$), confirming sense-blind neurons spread
edits indiscriminately across senses.

\section{Discussion}
\label{sec:discussion}

\paragraph{Are sense-blind neurons doing something useful?}
Neurons firing for all senses likely encode a pre-disambiguation
representation, and we do not dispute this. Our claim is narrower: these
neurons should not be counted as evidence of \emph{superposition}, because
they encode word form, not compressed concepts.
Causally, they have minimal impact on sense processing (probe accuracy
42\%, KL ratio $0.25\times$) while encoding word identity at 85\%.
Three independent lines rule out circularity: form detectors identified
\emph{without} sense labels overlap 67\% with sense-blind neurons; probing
shows a clean dissociation independent of the decomposition; and causal
ablation of sense-blind neurons has minimal effect on sense processing.

\paragraph{Implications for SAEs and practice.}
SAE evaluation assumes features are monosemantic~\citep{bricken2023monosemanticity, templeton2024scaling}, but 18--36\% of features active for polysemous words are monosemantic for a \emph{word form}, not a \emph{concept}---as illustrated by feature \#5132 (Section~\ref{sec:results-sae}), labeled ``intensifying adjectives'' on Neuronpedia despite firing sense-blind for 232 unrelated words. Since polysemous words are over 80\% of frequent vocabulary~\citep{rodd2002making}, such features are likely common. When practitioners steer or edit using SAE features~\citep{arditi2024refusal, meng2022locating}, sense-blind features spread changes across senses; our classification provides a filter.

\paragraph{Recommendations.}
For SAE evaluation: check whether a feature's top-activating examples share a word form, and test sense discrimination if so. For model editing: classify target neurons by SSI and restrict edits to sense-selective neurons. For polysemanticity measurement: report $R_\text{lex}$ or apply the LIS projection before computing overlap metrics. Looking ahead, a lexical-identity penalty in SAE training could produce sense-level monosemantic features.

\paragraph{Limitations.}
Absolute ablation effects are small ($<$3~ppl points), as expected from $\sim$150 of 3k--20k neurons; the primary evidence is the condition ordering itself, not the downstream magnitudes.
SemCor's vintage text is validated on modern Wikipedia (Appendix~\ref{app:modern}), and the ordering replicates across nine models on four corpora. The 57\% confound ratio applies only to polysemous inputs.
The corrected score requires sense labels; the LIS removal is more deployable but a fully unsupervised correction remains open (Appendix~\ref{app:unsupervised_lis}).

\section{Conclusion}
\label{sec:conclusion}

A substantial portion of what current metrics call polysemanticity is lexical identity, not superposition. Using a factorial decomposition across nine transformers (110M--70B), we showed that sharing a word form consistently produces more neuron overlap than sharing a meaning. The confound propagates into SAEs---where individual features fire sense-blind for hundreds of words---occupies a compact removable subspace, and causes collateral damage when editing or steering model behavior.

These findings do not deny that genuine superposition exists. They do show that the lexical signal must be separated out before the remaining overlap can be attributed to compression. We hope the tools introduced here---$R_\text{lex}$, the sense-selective/blind classification, and the lexical identity subspace---provide a starting point for polysemanticity measurements that distinguish word form from concept.

\bibliography{colm2026_conference}
\bibliographystyle{colm2026_conference}

\newpage
\appendix

\section*{Appendix: Table of Contents}

\noindent Click any entry to jump directly to that section.

\vspace{4mm}
\small
\renewcommand{\arraystretch}{1.4}
\begin{tabular}{@{}p{0.07\columnwidth}p{0.80\columnwidth}r@{}}

\hyperref[app:dataset]{\textbf{App.~\ref*{app:dataset}}} &
\hyperref[app:dataset]{\textbf{Dataset Details}} & \hyperref[app:dataset]{p.\,\pageref*{app:dataset}} \\
& \hyperref[app:dataset]{\quad$\triangleright$ Source corpus and selection criteria (SemCor, 407 polysemous words)} & \\
& \hyperref[app:dataset]{\quad$\triangleright$ Cleaning pipeline: function verb exclusion, deduplication, length filter} & \\
& \hyperref[app:dataset]{\quad$\triangleright$ Sense distance verification (Wu-Palmer similarity thresholds)} & \\
& \hyperref[app:dataset]{\quad$\triangleright$ Condition pair counts (Table~\ref*{tab:pair_counts}) and POS distribution} & \\[6pt]

\hyperref[app:results]{\textbf{App.~\ref*{app:results}}} &
\hyperref[app:results]{\textbf{Full Numerical Results}} & \hyperref[app:results]{p.\,\pageref*{app:results}} \\
& \hyperref[app:rlex]{\quad$\triangleright$ $R_\text{lex}$ by layer for all models (Tables~\ref*{tab:rlex_all},~\ref*{tab:rlex_large})} & \\
& \hyperref[app:embedding]{\quad$\triangleright$ Token-embedding baseline comparison across 8 models (Table~\ref*{tab:emb_baseline})} & \\
& \hyperref[app:ssi]{\quad$\triangleright$ Sense selectivity index: $\SSI > 2$ fractions per layer, threshold sensitivity (Table~\ref*{tab:ssi_all})} & \\
& \hyperref[app:sae]{\quad$\triangleright$ SAE collision: 32k vs.\ 128k dictionary comparison (Table~\ref*{tab:sae_compare})} & \\
& \hyperref[app:syn_validation]{\quad$\triangleright$ Synonym subset validation, POS stratification, Mann-Whitney tests (Table~\ref*{tab:syn_validation})} & \\[6pt]

\hyperref[app:figures]{\textbf{App.~\ref*{app:figures}}} &
\hyperref[app:figures]{\textbf{Additional Figures}} & \hyperref[app:figures]{p.\,\pageref*{app:figures}} \\
& \hyperref[fig:unit_overlap]{\quad$\triangleright$ Jaccard overlap of active-neuron sets across layers (GPT-2)} & \\
& \hyperref[fig:per_word_full]{\quad$\triangleright$ Per-word cross-sense cosine similarity heatmap} & \\
& \hyperref[fig:rlex_no_syn]{\quad$\triangleright$ $R_\text{lex}$ vs.\ $R_\text{lex}^\text{no-syn}$ comparison} & \\
& \hyperref[fig:mag_div_app]{\quad$\triangleright$ Magnitude divergence across layers} & \\
& \hyperref[fig:similarity_app]{\quad$\triangleright$ Condition means at representative layers} & \\
& \hyperref[fig:rlex_app]{\quad$\triangleright$ $R_\text{lex}$ across all 9 models with 95\% bootstrap CIs} & \\[6pt]

\hyperref[app:supplementary]{\textbf{App.~\ref*{app:supplementary}}} &
\hyperref[app:supplementary]{\textbf{Supplementary Analyses}} & \hyperref[app:supplementary]{p.\,\pageref*{app:supplementary}} \\
& \hyperref[app:supplementary]{\quad$\triangleright$ Sense accuracy after ablation (causal validation)} & \\
& \hyperref[app:supplementary]{\quad$\triangleright$ Concrete causal example: \emph{plant} (factory vs.\ botanical)} & \\
& \hyperref[app:supplementary]{\quad$\triangleright$ Cross-model scaling details (Pythia-12B, LLaMA-2-13B, LLaMA-2-70B)} & \\
& \hyperref[app:supplementary]{\quad$\triangleright$ The U-shaped trajectory: analysis and interpretation} & \\[6pt]

\hyperref[app:attention]{\textbf{App.~\ref*{app:attention}}} &
\hyperref[app:attention]{\textbf{Attention Representations}} & \hyperref[app:attention]{p.\,\pageref*{app:attention}} \\
& \hyperref[app:attention]{\quad$\triangleright$ Analysis of attention-head outputs (complementary to MLP analysis)} & \\[6pt]

\hyperref[app:reproducibility]{\textbf{App.~\ref*{app:reproducibility}}} &
\hyperref[app:reproducibility]{\textbf{Reproducibility \& Validation}} & \hyperref[app:reproducibility]{p.\,\pageref*{app:reproducibility}} \\
& \hyperref[app:reproducibility]{\quad$\triangleright$ Compute requirements (GPU hours, hardware)} & \\
& \hyperref[app:reproducibility]{\quad$\triangleright$ Software versions and dependencies} & \\
& \hyperref[app:reproducibility]{\quad$\triangleright$ Formal metric definitions (cosine similarity, Jaccard overlap, magnitude divergence, SSI)} & \\
& \hyperref[app:modern]{\quad$\triangleright$ Modern corpus validation on CoarseWSD-20 (Wikipedia sentences, Table~\ref*{tab:modern_validation})} & \\

\end{tabular}
\renewcommand{\arraystretch}{1.0}
\normalsize

\newpage

\section{Dataset details}
\label{app:dataset}

\paragraph{Source corpus and selection.}
All sentences are drawn from SemCor~\citep{miller1993semantic}, a
sense-tagged corpus built on the Brown Corpus with human-annotated WordNet
sense labels. From the full corpus (37{,}176 sentences, 88{,}334 sense-tagged
tokens), we identified 494 content words (nouns and verbs) with at least two
senses, each attested in $\geq 8$ sentences.

\paragraph{Cleaning pipeline.}
Three cleaning steps were applied in order:
\begin{enumerate}
  \item \textbf{Function verb exclusion.} We removed 31 high-frequency
    function verbs (\emph{be, have, do, go, make, get, take, come, give, say},
    etc.)\ whose multiple WordNet ``senses'' reflect grammatical rather than
    semantic distinctions.
  \item \textbf{Cross-word deduplication.} 3{,}591 sentences that appear
    under multiple words in SemCor were kept only for their first word,
    preventing cross-word contamination.
  \item \textbf{Length filter.} Sentences shorter than 30 characters were
    removed (typically sentence fragments or captions).
\end{enumerate}
After cleaning, 407 words remain with $\geq 5$ sentences per sense (mean:
10.3 sentences per sense; range: 5--15).

\paragraph{Sense distance verification.}
We computed Wu-Palmer similarity~\citep{wu1994verb} between the two selected
WordNet synsets for each word. The mean similarity across all 407 word pairs is
$0.16$ (range: $0.10$--$0.22$). We enforced a maximum threshold of $0.50$ to
exclude near-synonymous senses.

\paragraph{Condition pair counts.}
Table~\ref{tab:pair_counts} shows the number of sentence pairs constructed for
each condition. SL and PS pairs are available for all 407 words. The SYN
condition requires a WordNet synonym attested in SemCor with $\geq 3$
sentences; 172 words have such a synonym for sense~A and 160 for sense~B. CL
pairs are formed by randomly sampling sentences from unrelated words, excluding
the current word's own sentences.

\begin{table}[h]
\begin{center}
\small
\begin{tabular}{lrrr}
\toprule
\textbf{Condition} & \textbf{Words} & \textbf{Pairs (total)} & \textbf{Pairs/word (mean)} \\
\midrule
Same-Lemma (SL)    & 407 & $\sim$40{,}700 & 100 \\
Polysemantic (PS)  & 407 & $\sim$40{,}700 & 100 \\
Synonym (SYN)      & 251 & $\sim$12{,}500 & 50  \\
Cross-Lemma (CL)   & 407 & $\sim$40{,}700 & 100 \\
\bottomrule
\end{tabular}
\end{center}
\caption{Sentence pair counts by condition. SYN coverage is lower because
synonyms must be attested in SemCor. All pairs are capped at 200 per word.}
\label{tab:pair_counts}
\end{table}

\paragraph{POS distribution.}
Of the 407 words, 156 (38\%) are nouns and 251 (62\%) are verbs. 13 lemmas
appear twice with different POS tags (e.g., \emph{brush} as noun and verb),
which we treat as separate entries since they have distinct sense pairs.
The full word list, per-word SSI distributions, and SAE feature-level
collision data (including the 232 words for which feature \#5132 is
sense-blind) are released with our code.

\section{Full numerical results}
\label{app:results}

\subsection{$R_\text{lex}$ by layer}
\label{app:rlex}

Table~\ref{tab:rlex_all} reports $R_\text{lex}$ computed from cosine similarity
at each layer for the 12-layer models. $R_\text{lex}$ decreases with depth in
all models, with BERT and ELECTRA achieving lower minima than GPT-2, reflecting
stronger sense separation in bidirectional models.

\begin{table}[h]
\begin{center}
\small
\begin{tabular}{rcccc}
\toprule
\textbf{Layer} & \textbf{GPT-2} & \textbf{BERT} & \textbf{ELECTRA} & \textbf{GPT-2-Med} \\
\midrule
0  & .735 & .662 & .686 & .751 \\
1  & .629 & .566 & .528 & .700 \\
2  & .560 & .552 & .482 & .690 \\
3  & .494 & .471 & .428 & .612 \\
4  & .446 & .398 & .367 & .549 \\
5  & .441 & .409 & .307 & .524 \\
6  & .423 & .367 & .250 & .521 \\
7  & .418 & .313 & .268 & .498 \\
8  & .447 & .275 & .264 & .443 \\
9  & .462 & .225 & .309 & .454 \\
10 & .399 & .360 & .381 & .440 \\
11 & .290 & .209 & .542 & .429 \\
\midrule
\textbf{Min} & .290 & .209 & .250 & .270 \\
\bottomrule
\end{tabular}
\end{center}
\caption{$R_\text{lex}$ (cosine-based) by layer for 12-layer models and
GPT-2-Medium (first 12 of 24 layers shown). GPT-2-Medium's minimum of $0.27$
occurs at layer~23. BERT achieves its minimum at layer~9 ($0.23$), ELECTRA at
layer~6 ($0.25$). GPT-2's $R_\text{lex}$ remains above $0.29$ throughout.}
\label{tab:rlex_all}
\end{table}

\begin{table}[h]
\begin{center}
\small
\begin{tabular}{rcc}
\toprule
\textbf{Layer} & \textbf{Pythia-12B} & \textbf{LLaMA-2-13B} \\
\midrule
0  & .913 & .911 \\
5  & .516 & .406 \\
10 & .377 & .262 \\
15 & .360 & .456 \\
20 & .422 & .476 \\
25 & .391 & .348 \\
30 & .440 & .335 \\
35 & .288 & .281 \\
39 & --- & .261 \\
\midrule
\textbf{Min} & .288 & .250 \\
\bottomrule
\end{tabular}
\end{center}
\caption{$R_\text{lex}$ (cosine-based) at representative layers for the larger models.
Pythia-12B has 36 layers; LLaMA-2-13B has 40 layers. Both start with $R_\text{lex} > 0.9$
and decline with depth, reaching minima comparable to the smaller models.}
\label{tab:rlex_large}
\end{table}

\subsection{Token-embedding baseline}
\label{app:embedding}

Table~\ref{tab:emb_baseline} reports the embedding-only $R_\text{lex}^\text{emb}$
for each model, computed from raw token embeddings before any transformer
computation. MLP-level $R_\text{lex}$ exceeds this baseline at layer~0 in all
autoregressive models, confirming that the confound is amplified beyond trivial
token reuse.

\begin{table}[h]
\begin{center}
\small
\begin{tabular}{lccc}
\toprule
\textbf{Model} & $R_\text{lex}^\text{emb}$ & $R_\text{lex}$ (L0) & \textbf{Exceeds?} \\
\midrule
GPT-2        & 0.713 & 0.735 & Yes \\
GPT-2-Med    & 0.720 & 0.751 & Yes \\
BERT-base    & 0.778 & 0.662 & No  \\
ELECTRA-base & 0.727 & 0.686 & No  \\
Pythia-1B    & 0.821 & 0.836 & Yes \\
Pythia-6.9B  & 0.882 & 0.910 & Yes \\
Pythia-12B   & 0.891 & 0.913 & Yes \\
LLaMA-2-13B & 0.815 & 0.911 & Yes \\
\bottomrule
\end{tabular}
\end{center}
\caption{Token-embedding baseline comparison. All autoregressive models show
MLP-level $R_\text{lex}$ exceeding the embedding baseline at layer~0, confirming
non-trivial amplification. Bidirectional models (BERT, ELECTRA) show MLP
$R_\text{lex}$ below the embedding baseline at layer~0, suggesting early-layer
bidirectional processing partially resolves lexical overlap---but the confound
nonetheless persists at all layers.}
\label{tab:emb_baseline}
\end{table}

\subsection{Sense selectivity index}
\label{app:ssi}

Table~\ref{tab:ssi_all} reports the fraction of MLP neurons with $\SSI > 2$
(sense-selective) at each layer for the 12-layer models. In all models, fewer
than 1\% of neurons are sense-selective at any layer.
The choice of threshold ($\SSI > 2$) follows the standard Cohen's $d$
convention for a large effect. Relaxing to $\SSI > 1.5$ roughly doubles the
count but does not change the qualitative pattern: sense-selective neurons
remain a small minority ($<$2\%) and the probing dissociation (sense accuracy
for blind neurons near chance, selective neurons $>$90\%) holds across
threshold choices. The sense-blind threshold ($\SSI < 0.5$) similarly follows
the small-effect convention; using $\SSI < 0.3$ or $\SSI < 0.8$ shifts counts
but preserves the functional split.

\begin{table}[h]
\begin{center}
\small
\begin{tabular}{rcccc}
\toprule
\textbf{Layer} & \textbf{GPT-2} & \textbf{BERT} & \textbf{ELECTRA} & \textbf{GPT-2-Med} \\
\midrule
0  & 0.88 & 0.63 & 0.57 & 0.70 \\
1  & 0.77 & 0.78 & 0.64 & 0.75 \\
2  & 0.83 & 0.64 & 0.76 & 0.80 \\
3  & 0.84 & 0.61 & 0.74 & 0.85 \\
4  & 0.72 & 0.67 & 0.81 & 0.78 \\
5  & 0.74 & 0.63 & 0.83 & 0.74 \\
6  & 0.74 & 0.62 & 0.75 & 0.71 \\
7  & 0.78 & 0.60 & 0.69 & 0.73 \\
8  & 0.78 & 0.66 & 0.63 & 0.68 \\
9  & 0.77 & 0.78 & 0.54 & 0.73 \\
10 & 0.75 & 0.88 & 0.53 & 0.71 \\
11 & 0.72 & 0.79 & 0.49 & 0.73 \\
\bottomrule
\end{tabular}
\end{center}
\caption{Fraction (\%) of neurons with $\SSI > 2$ at each layer.  The
denominator is $d_\text{mlp}$ (3072 for GPT-2/BERT/ELECTRA, 4096 for
GPT-2-Med). Pythia-1B's layer~0 has the highest fraction (1.80\%) due to its
larger MLP producing more sense-differentiating neurons in absolute terms,
but the fraction remains very small.}
\label{tab:ssi_all}
\end{table}

\subsection{SAE collision: 32k vs.\ 128k dictionary comparison}
\label{app:sae}

Table~\ref{tab:sae_compare} compares the two SAE dictionary sizes across all
five probed layers. The collision ratios are strikingly similar despite a
$4\times$ difference in dictionary capacity, confirming that the lexical
confound is not resolved by increasing the number of SAE features.

\begin{table}[h]
\begin{center}
\small
\begin{tabular}{rcccccc}
\toprule
 & \multicolumn{3}{c}{\textbf{32k SAE}} & \multicolumn{3}{c}{\textbf{128k SAE}} \\
\cmidrule(lr){2-4} \cmidrule(lr){5-7}
\textbf{Layer} & Active & Blind & Coll.\% & Active & Blind & Coll.\% \\
\midrule
0  & 33.6 & 10.4 & 31.7 & 31.5 & 9.8 & 32.6 \\
2  & 27.8 &  7.4 & 29.1 & 26.4 & 7.6 & 31.3 \\
5  & 21.3 &  4.3 & 24.1 & 20.0 & 4.4 & 25.7 \\
8  & 18.6 &  2.7 & 17.7 & 15.2 & 2.6 & 22.4 \\
11 & 18.4 &  3.9 & 28.9 & 15.2 & 2.9 & 25.9 \\
\bottomrule
\end{tabular}
\end{center}
\caption{SAE collision analysis: 32k vs.\ 128k features. \emph{Active}: mean
features active per word ($>$30\% firing rate in either sense).
\emph{Blind}: mean sense-blind features (Cohen's $d < 0.5$, fires for both
senses). \emph{Coll.\%}: collision ratio (blind/active). Both SAEs show
similar collision ratios at every layer, confirming that increasing dictionary
size does not resolve the confound.}
\label{tab:sae_compare}
\end{table}

\subsection{Synonym subset validation}
\label{app:syn_validation}

The full $R_\text{lex}$ requires the SYN condition, which is available for 251
of 407 words (62\%).  We verify that this subset is unbiased by testing whether
the core lexical confound---the per-word SL--PS cosine gap---differs between
synonym-available and synonym-absent words.

\paragraph{Subset comparisons.}
Table~\ref{tab:syn_validation} reports Mann-Whitney $U$ tests at five
representative layers in each model.  No test reaches significance
($p < 0.05$) in any model at any layer; most $p$-values exceed $0.2$.
This confirms that the two subsets are statistically indistinguishable on
the metric that drives $R_\text{lex}$.

\begin{table}[h]
\begin{center}
\small
\setlength{\tabcolsep}{3pt}
\begin{tabular}{lcccccc}
\toprule
\textbf{Model} & \textbf{L0} & \textbf{L$\frac{n}{4}$} & \textbf{L$\frac{n}{2}$} & \textbf{L$\frac{3n}{4}$} & \textbf{L$_{n-1}$} & \textbf{min $p$} \\
\midrule
GPT-2       & .43 & .34 & .45 & .56 & .37 & .28 \\
GPT-2-Med   & .50 & .28 & .38 & .42 & .28 & .28 \\
BERT-base   & .57 & .31 & .14 & .18 & .16 & .14 \\
ELECTRA     & .48 & .33 & .50 & .53 & .26 & .26 \\
Pythia-1B   & .62 & .20 & .33 & .30 & .16 & .16 \\
Pythia-6.9B & .80 & .43 & .26 & .13 & .25 & .13 \\
\bottomrule
\end{tabular}
\end{center}
\caption{Mann-Whitney $U$ test $p$-values comparing the per-word SL--PS
cosine gap between synonym-available ($n = 251$) and synonym-absent ($n = 156$)
words.  No test reaches significance; the minimum $p$-value across all models
and layers is $0.13$ (Pythia-6.9B, layer~24).}
\label{tab:syn_validation}
\end{table}

\paragraph{POS stratification.}
The synonym subset is POS-imbalanced: 69\% verbs vs.\ 50\% in the
non-synonym set ($\chi^2 = 13.8$, $p < 0.001$).  Nouns exhibit a larger
SL--PS gap than verbs ($p < 0.005$ at all layers in GPT-2), reflecting
stronger lexical identity effects for nouns.  However, both POS categories
show a consistently positive SL--PS gap at every layer in every model,
confirming that the confound is universal; the POS imbalance affects magnitude
but not the direction or existence of the effect.

\paragraph{Other properties.}
Wu-Palmer similarity distributions are indistinguishable between subsets
(KS $p = 0.98$, Mann-Whitney $p = 0.91$).  Sentence counts per word are
likewise non-significant (KS $p = 0.68$).  Word length shows a small
difference (6.3 vs.\ 5.9 characters; KS $p < 0.001$), but this does not
correlate with the SL--PS gap at any layer.

\section{Additional figures}
\label{app:figures}

\begin{figure}[h]
\centering
\includegraphics[width=\columnwidth]{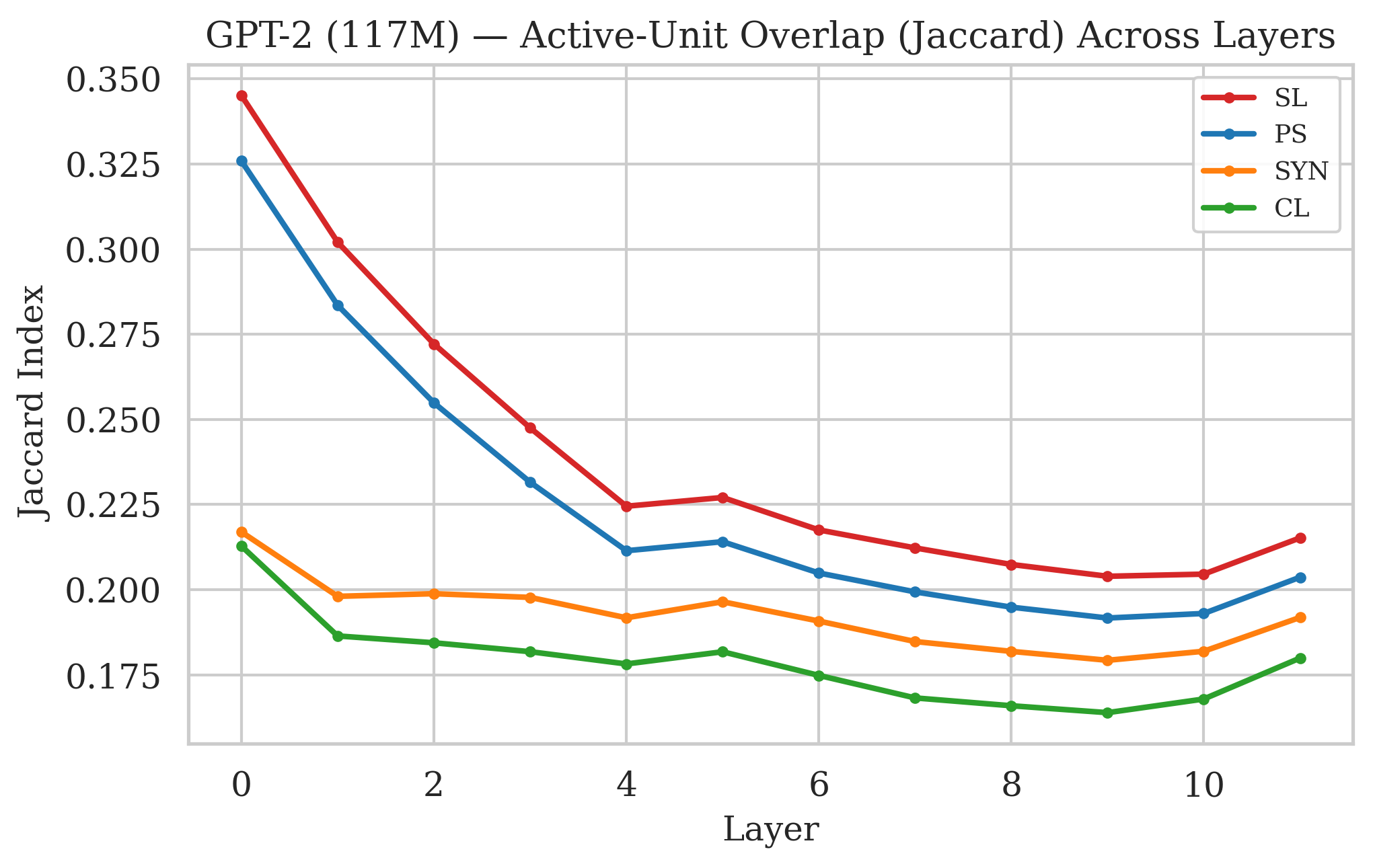}
\caption{Jaccard overlap of active-neuron sets across layers (GPT-2).
The condition ordering $\SL > \PS > \text{SYN} > \CL$ is consistent with the
cosine similarity results, confirming that the lexical confound affects which
neurons fire, not just how strongly they fire.}
\label{fig:unit_overlap}
\end{figure}

\begin{figure}[h]
\centering
\includegraphics[width=\columnwidth]{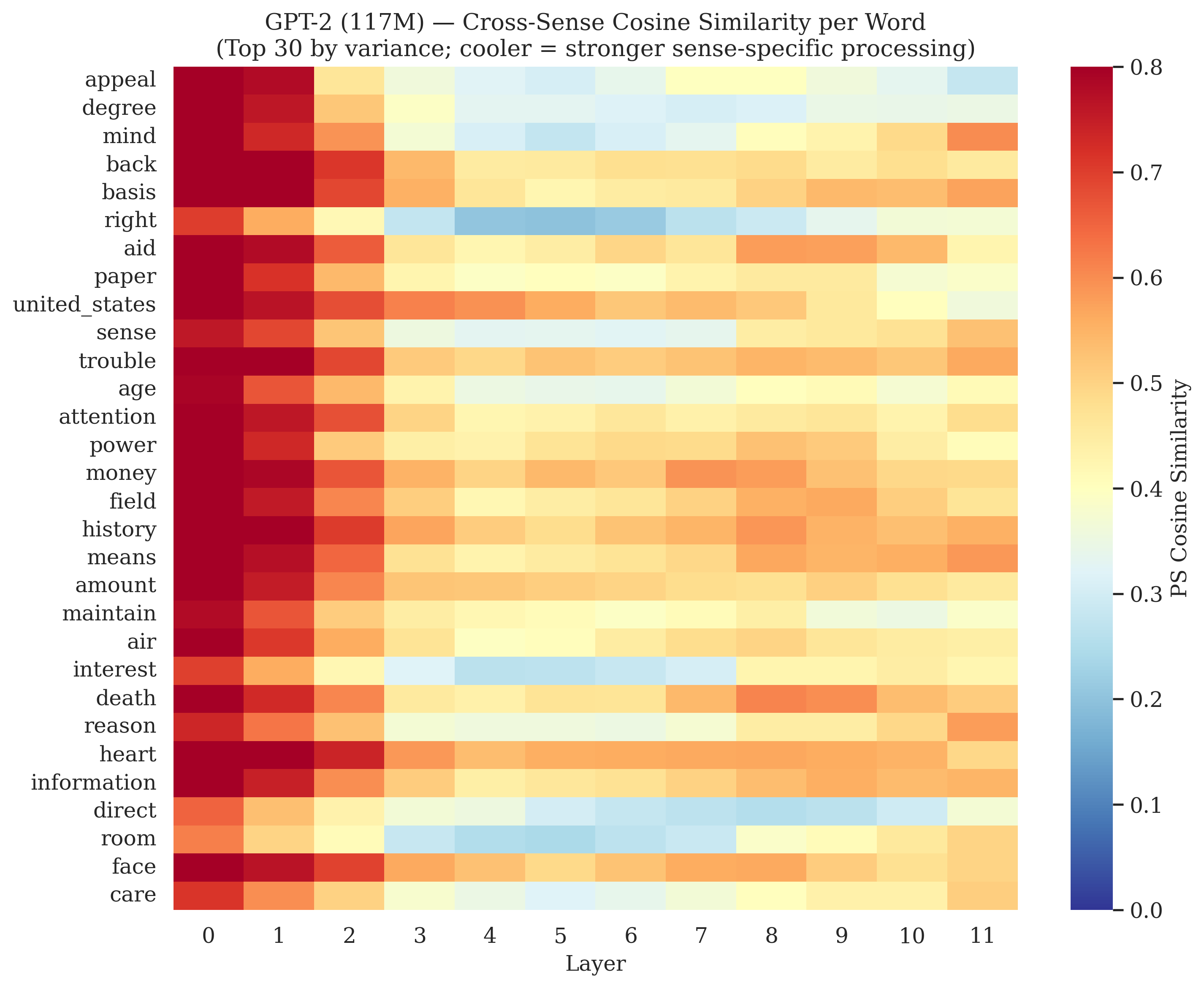}
\caption{Cross-sense (PS) cosine similarity per word per layer in GPT-2
(top 30 words by variance).  Cooler colors indicate lower similarity (stronger
sense-specific processing).  Most words exhibit a U-shaped trajectory with a
dip in mid-layers, though the depth and timing of the minimum varies
substantially across words---sense separation is word-specific, not driven by
a shared mechanism.}
\label{fig:per_word_full}
\end{figure}

\begin{figure}[h]
\centering
\includegraphics[width=\columnwidth]{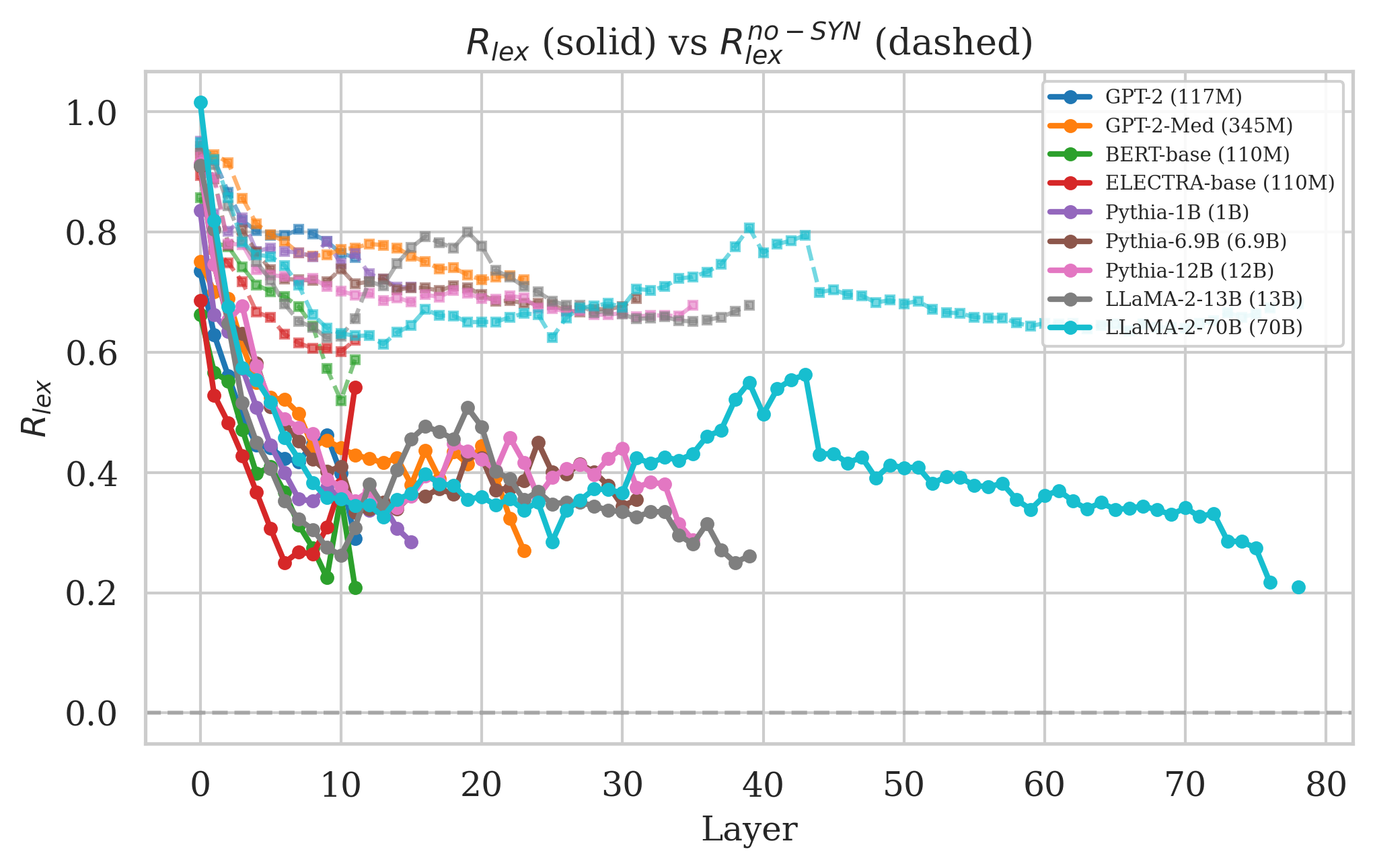}
\caption{$R_\text{lex}$ (solid) vs.\ $R_\text{lex}^\text{no-syn}$ (dashed)
for all eight models.  The alternative metric
$R_\text{lex}^\text{no-syn} = (\bar{M}_\PS - \bar{M}_\CL) /
(\bar{M}_\SL - \bar{M}_\CL)$ does not require synonym data and can be
computed for all 407 words.  The qualitatively identical trajectories confirm
that our findings do not depend on synonym coverage (62\% of words, validated
by subset invariance test; see Appendix~\ref{app:syn_validation}).}
\label{fig:rlex_no_syn}
\end{figure}

\begin{figure}[h]
\centering
\includegraphics[width=\columnwidth]{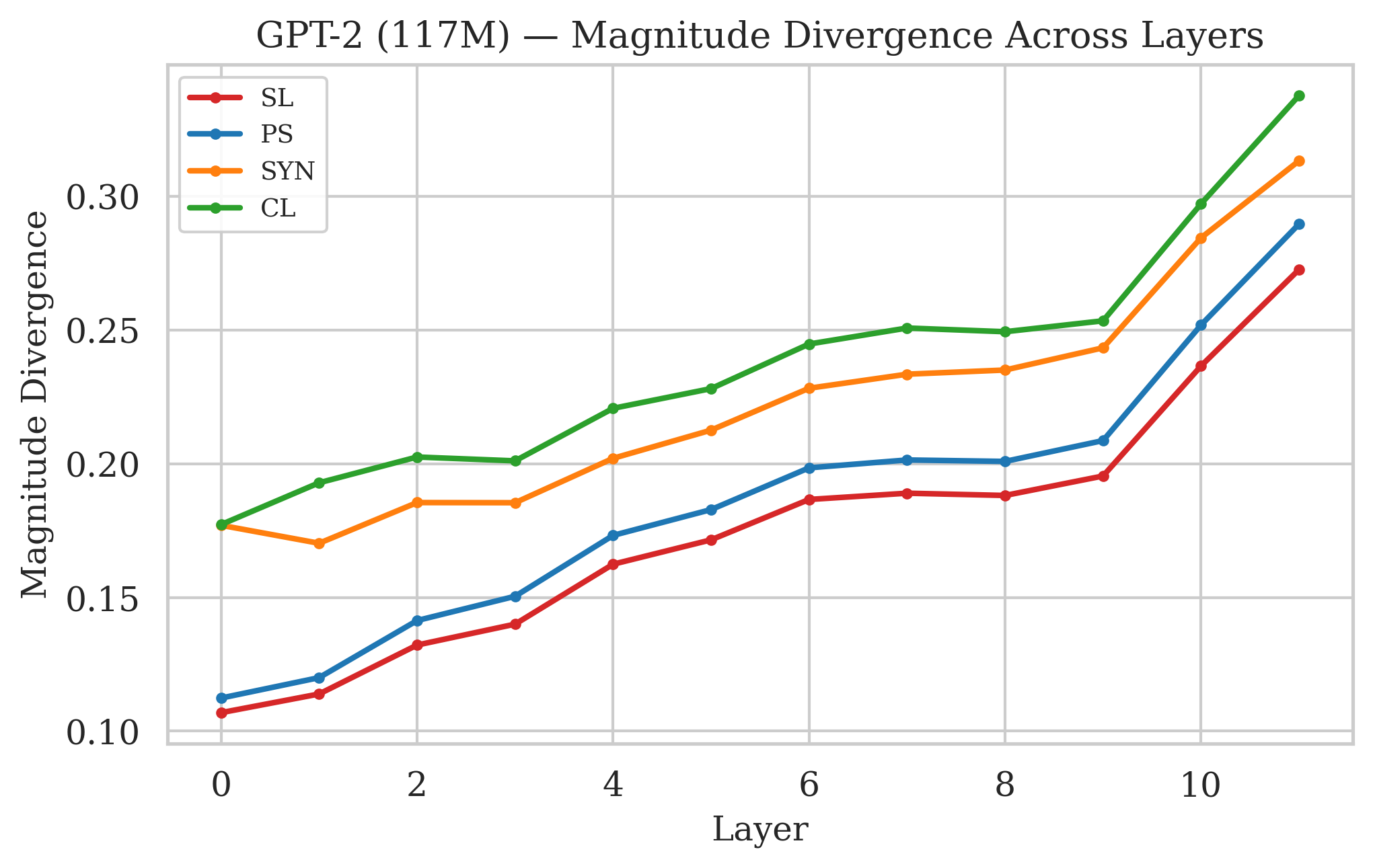}
\caption{Mean magnitude divergence ($\Delta = \frac{1}{|A \cap B|}
\sum_{i \in A \cap B} |a_i - b_i|$) by condition and layer.  Unlike cosine
similarity (which shows a U-shaped trajectory), magnitude divergence increases
monotonically with depth, reflecting growing activation scale differences as
representations become more specialized.  The condition ordering is
reversed---$\CL > \text{SYN} > \PS > \SL$---because more dissimilar conditions
produce larger activation differences among their shared neurons.}
\label{fig:mag_div_app}
\end{figure}

\begin{figure}[h]
\centering
\includegraphics[width=\columnwidth]{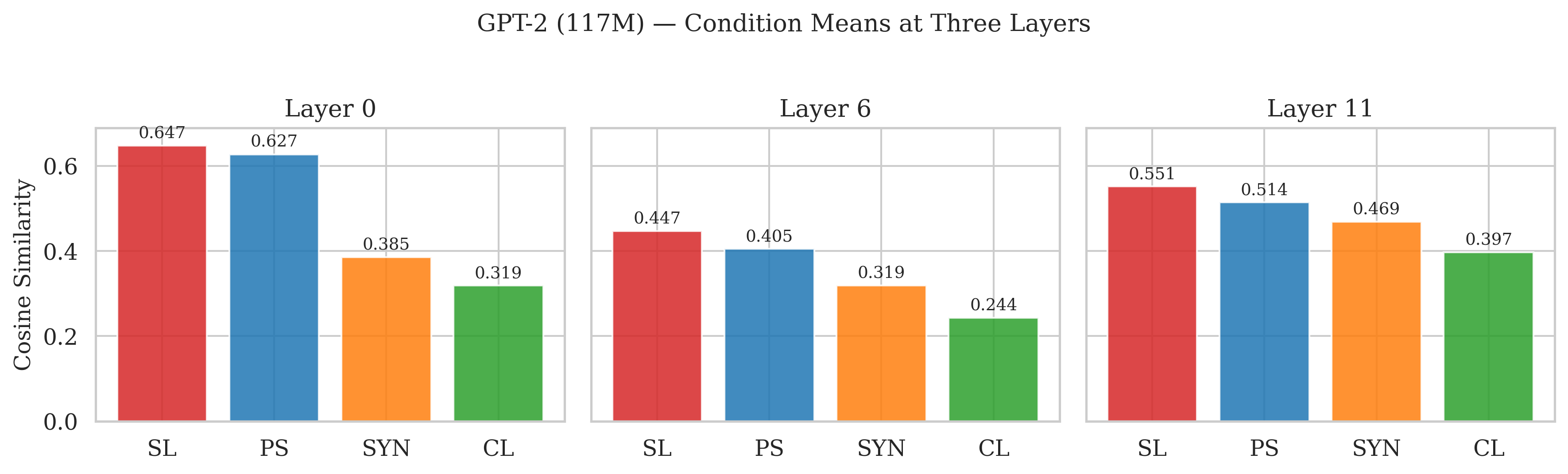}
\caption{Condition means at three representative layers in GPT-2 (layers~0, 6,
11).  At layer~0, \PS{} and \SL{} are close while SYN and \CL{} are distant.
By layer~6, the gap between \PS{} and \SL{} widens.  At layer~11, all
conditions retain the same ordering with moderate separation.}
\label{fig:similarity_app}
\end{figure}

\begin{figure}[h]
\centering
\includegraphics[width=\columnwidth]{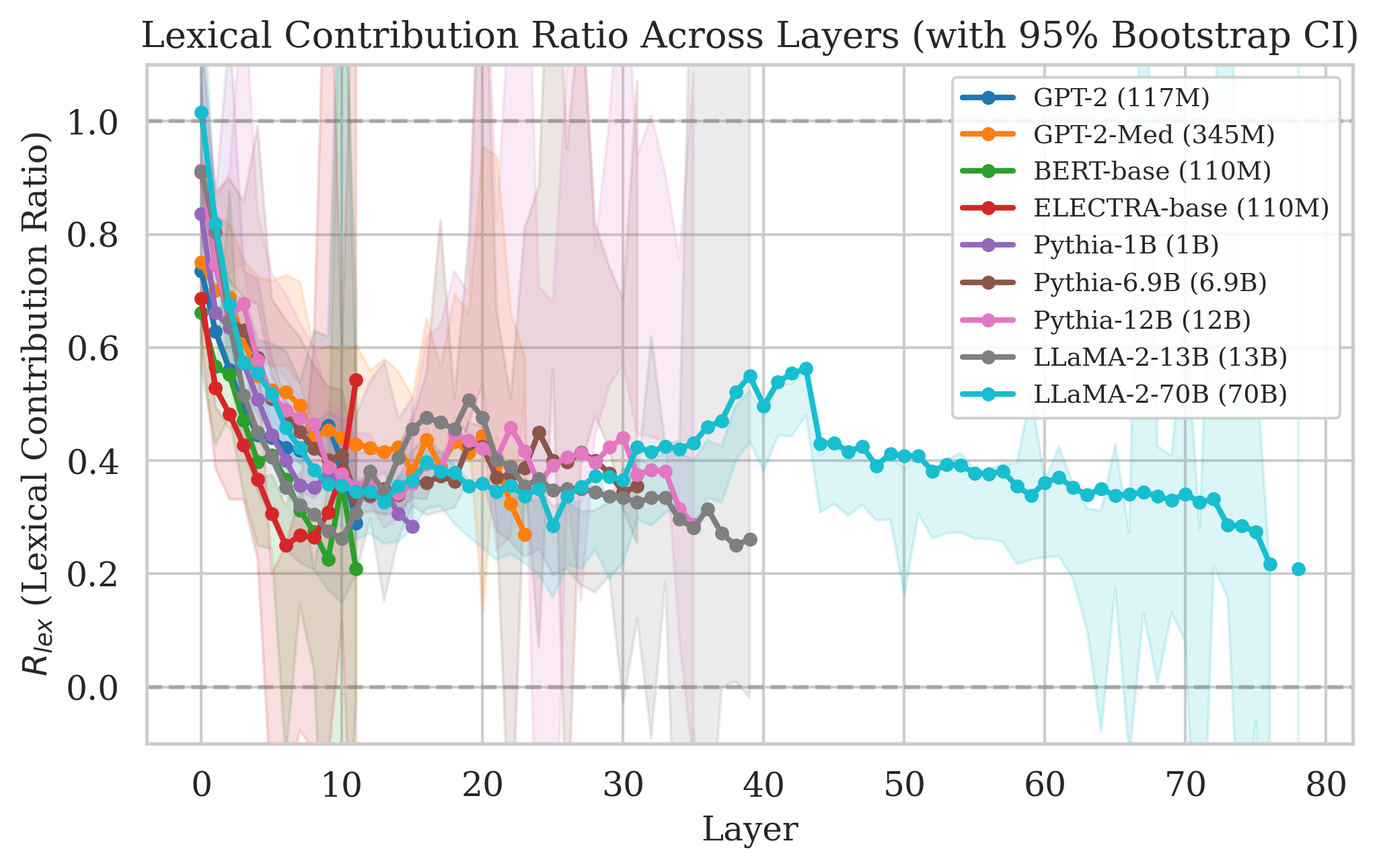}
\caption{$R_\text{lex}$ (lexical contribution ratio) across layers for all nine models
with 95\% bootstrap CIs. The lexical contribution decreases with depth but
never reaches zero in any model. Larger models (LLaMA-2-70B, cyan) show sustained
$R_\text{lex} \approx 0.35$--$0.45$ across 80 layers.}
\label{fig:rlex_app}
\end{figure}

\section{Supplementary analyses}
\label{app:supplementary}

\paragraph{Sense accuracy after ablation.}
On GPT-2, clean (no ablation) sense accuracy averages 73\% (measured via
diagnostic token probabilities on held-out sentences).
After ablating sense-selective neurons, accuracy drops by 8 percentage points;
after ablating sense-blind neurons, it drops by $<$1~pp (indistinguishable
from random variation), confirming that sense-selective neurons are causally
necessary for the model's sense-discriminating behavior.

\paragraph{Concrete causal example.}
For the word \emph{bank}, sense-A (financial institution) and sense-B
(geographic land) yield distinct diagnostic tokens (\emph{loan}, \emph{account}
for sense~A; \emph{river}, \emph{shore} for sense~B).
Ablating sense-A-selective neurons increases perplexity on financial-institution
sentences while leaving river-bank sentences unchanged; ablating sense-blind
neurons has negligible impact on either.

\paragraph{Cross-model scaling details.}
Pythia-1B ($>$5$\times$ GPT-2's parameters) shows a qualitatively identical
trajectory.
Pythia-6.9B confirms the confound persists at the scale where modern
interpretability work typically operates.
Pythia-12B and LLaMA-2-13B (gated SiLU MLP, a fourth architecture family)
maintain the strict condition ordering at every layer; LLaMA-2-13B achieves
the lowest absolute similarities but the same pattern, with $R_\text{lex}$
ranging from $0.91$ (layer~0) to $0.25$ (layer~38).

\paragraph{The U-shaped trajectory.}
All conditions exhibit a U-shaped similarity trajectory---high in early layers,
lowest in mid-layers, partially recovering in final layers.
This is consistent with prior observations that late-layer representations
reconverge toward the output vocabulary
space~\citep{nostalgebraist2020logitlens, geva2022transformer}, possibly
partially undoing sense-specific processing.

\paragraph{LIS dose-response.}
\label{app:lis_dose}
Table~\ref{tab:lis_ablation} shows the full dose-response from projecting out
the top-$k$ LIS components. The PS--SYN gap closes monotonically; PS
similarity rises slightly (removing between-word variance makes within-word
representations more uniform) while SYN drops (shared meaning partly encoded
along word-form directions). The gap is the informative quantity.

\begin{table}[h]
\begin{center}
\small
\begin{tabular}{lcccc}
\toprule
& \textbf{PS--SYN gap} & $R_\text{lex}$ & $\Delta$\textbf{PS} & $\Delta$\textbf{SYN} \\
\midrule
\multicolumn{5}{c}{\textit{GPT-2 (d=3072, layer-averaged)}} \\
\midrule
Baseline         & 0.119 & 0.65 & ---  & --- \\
$k = 10$ (0.3\%) & 0.087 & 0.52 & $+$0.008 & $-$0.032 \\
$k = 20$ (0.7\%) & 0.047 & 0.41 & $+$0.019 & $-$0.054 \\
$k = 50$ (1.6\%) & $-$0.028 & $<$0  & $+$0.031 & $-$0.116 \\
\midrule
\multicolumn{5}{c}{\textit{Pythia-1B (d=8192, layer-averaged)}} \\
\midrule
Baseline         & 0.097 & 0.59 & ---  & --- \\
$k = 10$ (0.1\%) & 0.062 & 0.47 & $+$0.016 & $-$0.019 \\
$k = 20$ (0.2\%) & 0.033 & 0.29 & $+$0.025 & $-$0.040 \\
$k = 50$ (0.6\%) & $-$0.032 & $<$0 & $+$0.037 & $-$0.092 \\
\bottomrule
\end{tabular}
\end{center}
\caption{LIS dose-response ablation. Removing $k = 20$ components
($\le$\,1\% of dimensions) reduces $R_\text{lex}$ by 37--51\%.
At $k = 50$, the PS--SYN gap reverses---the confound is eliminated.}
\label{tab:lis_ablation}
\end{table}

\paragraph{Per-word collateral damage.}
The collateral-damage experiment (Section~\ref{sec:results-causal}) ablates
sense-blind vs.\ sense-selective neurons for individual words at layer~6.
Across 10 polysemous words tested, sense-blind ablation consistently
produces low specificity (both senses equally affected), while
sense-selective ablation is $5$--$72\times$ more targeted. The full
per-word results are included in the released code.

\paragraph{Unsupervised LIS detection.}
\label{app:unsupervised_lis}
To test whether the LIS can be approximated without sense annotations,
we replace WordNet synonyms with nearest neighbors from GPT-2's embedding
layer ($n = 393$ pairs). The resulting subspace overlaps with the
WordNet-based LIS at $0.19$--$0.25$ ($7$--$10\times$ above a random baseline
of $0.026$). This partial recovery suggests the subspace is identifiable
without labels, though a fully unsupervised method remains open.

\section{Attention representations}
\label{app:attention}

We apply the same four-condition decomposition to attention head outputs
(concatenated per-head output vectors at the target word position).
The ordering $\SL > \PS > \text{SYN} > \CL$ holds in attention
representations at every layer in all models, with $R_\text{lex}$ values
comparable to MLP values in early layers.
Since the residual stream sums attention and MLP outputs, the confound
propagates to SAEs trained on \texttt{resid\_post} (Section~\ref{sec:results-sae}).

\section{Reproducibility details}
\label{app:reproducibility}

\paragraph{Compute.}
All experiments were run on a university HPC cluster.
Models up to 1B parameters used NVIDIA V100 GPUs (16\,GB VRAM);
Pythia-6.9B required an A100 GPU (40\,GB VRAM);
Pythia-12B and LLaMA-2-13B each required an NVIDIA GH200 GPU (96\,GB VRAM).
Decomposition experiments took approximately 2--4 hours per model on a single
GPU (6--8 hours for the 12B and 13B models). The SAE collision experiments took approximately 20 minutes (32k SAE) and
30 minutes (128k SAE) on a V100.

\paragraph{Software.}
Models were loaded using HuggingFace Transformers (v4.36+) in
\texttt{float32} precision. SAE experiments used TransformerLens (v1.6+) and
SAE Lens (v3.0+). Random seed was fixed at 42 for all experiments. Bootstrap
confidence intervals used 10{,}000 resamples stratified by word.

\paragraph{Metric definitions (formal).}
For completeness, we provide the formal definitions of all metrics.

\emph{Cosine similarity:}
$\text{cos}(\mathbf{a}, \mathbf{b}) = \frac{\mathbf{a} \cdot \mathbf{b}}
{\|\mathbf{a}\|\;\|\mathbf{b}\|}$

\emph{Jaccard overlap:} A neuron is active if its absolute activation exceeds
the per-sentence median.
$J = |A \cap B| / |A \cup B|$.

\emph{Magnitude divergence:}
$\Delta = \frac{1}{|A \cap B|} \sum_{i \in A \cap B} |a_i - b_i|$

\emph{Sense Selectivity Index:}
$\SSI_i = |\mu_i^{(A)} - \mu_i^{(B)}| / s_i^\text{pooled}$
where $s_i^\text{pooled} = \sqrt{((n_A - 1)s_A^2 + (n_B - 1)s_B^2) /
(n_A + n_B - 2)}$. Neurons with $\SSI > 2$ are classified as sense-selective.

\subsection{Modern corpus validation (CoarseWSD-20)}
\label{app:modern}

A potential concern is that SemCor draws on the 1960s Brown Corpus.
To rule out the possibility that our findings are artifacts of vintage
text, we replicate the \emph{full} four-condition decomposition on
\textbf{CoarseWSD-20}~\citep{loureiro2021analysis}, a dataset of 20
polysemous nouns with sense-annotated \textbf{Wikipedia} sentences.
We construct all four conditions---SL, PS, SYN, and CL---by pairing
CoarseWSD-20 sense labels with WordNet synonyms whose sentences are
drawn from the same Wikipedia corpus (different words' entries),
ensuring the SYN condition uses modern text throughout.
We run the decomposition on GPT-2, BERT-base, and Pythia-1B.

\begin{table}[h]
\begin{center}
\small
\begin{tabular}{lccccl}
\toprule
\textbf{Model} & \textbf{SL} & \textbf{PS} & \textbf{SYN} & \textbf{CL} & \textbf{PS$>$SYN} \\
\midrule
GPT-2      & .90--.60 & .89--.47 & .31--.45 & .26--.45 & $p < .001$ $\checkmark$ \\
BERT-base  & .70--.58 & .54--.30 & .09--.32 & .05--.35 & $p < .01$ $\checkmark$ \\
Pythia-1B  & .99--.46 & .99--.27 & .09--.28 & .09--.26 & $p < .001$ $\checkmark$ \\
\bottomrule
\end{tabular}
\end{center}
\caption{Modern corpus validation on CoarseWSD-20 (Wikipedia sentences,
20 nouns). Ranges show cosine similarity across layers (layer~0 to final).
The critical PS $>$ SYN gap---indicating lexical form drives more overlap
than shared meaning---is statistically significant at every layer in all
three models (Wilcoxon signed-rank, Holm-Bonferroni corrected).
$R_\text{lex}$ bootstrap 95\% CI remains above zero everywhere.}
\label{tab:modern_validation}
\end{table}

The results replicate the SemCor findings on modern text.
The critical PS $>$ SYN gap is significant at every layer in all three
models, confirming that lexical form is a stronger driver of MLP neuron
overlap than semantic similarity even on Wikipedia text.
GPT-2's $R_\text{lex}$ ranges from $0.98$ (layer~0) to $0.57$
(layer~11), closely matching the SemCor trajectory ($0.74$--$0.29$).
Pythia-1B shows an even stronger confound ($R_\text{lex} \ge 0.41$
at all layers).
The full ordering $\SL > \PS > \text{SYN} > \CL$ holds at the
majority of layers; at a few late layers, SYN and CL converge (both
approach $0.3$--$0.4$), consistent with the SemCor pattern where all
conditions converge at depth.
The lexical confound is a property of how transformers represent
polysemous words, not an artifact of the SemCor corpus.

\end{document}